\documentclass{egpubl}
\usepackage{VMV2026}

 \WsPaper           

\usepackage[T1]{fontenc}
\usepackage{dfadobe}

\usepackage{cite}  
\BibtexOrBiblatex
\electronicVersion
\PrintedOrElectronic
\ifpdf \usepackage[pdftex]{graphicx} \pdfcompresslevel=9
\else \usepackage[dvips]{graphicx} \fi

\usepackage{egweblnk}

\usepackage{color}

\usepackage{amsmath}
\usepackage{amssymb}
\usepackage{booktabs}
\usepackage{float}
\usepackage{multirow}
\usepackage{url}
\usepackage{algorithm}
\usepackage{algpseudocode}
\usepackage{placeins}
\usepackage{longtable}
\usepackage{tabularx}
\usepackage{array}
\usepackage{adjustbox}

\newcolumntype{Y}{>{\centering\arraybackslash}X}

\usepackage{cleveref}
\crefname{figure}{Fig.}{Figs.}
\Crefname{figure}{Fig.}{Figs.}
\crefname{table}{Tab.}{Tabs.}
\Crefname{table}{Tab.}{Tabs.}
\crefname{section}{Sec.}{Secs.}
\Crefname{section}{Section}{Sections}
\crefname{equation}{Eq.}{Eqs.}
\Crefname{equation}{Eq.}{Eqs.}
\crefname{app}{App.}{Apps.}
\Crefname{app}{App.}{Apps.}

\usepackage{graphicx}
\usepackage{adjustbox}
\usepackage{tikz}   

\title[FittedGenes: Domain-Aware Sewing Pattern Layout Optimization]%
      {FittedGenes: Domain-Aware Sewing Pattern Layout Optimization}

\author[Barlas \& Korosteleva]{%
Ayşegül Barlas\\
ETH Zürich\\
\texttt{aysegulbarlas@gmail.com}
\and
Maria Korosteleva\\
Meshcapade GmbH\\
\texttt{masha@meshcapade.com}%
}

\begin{document}
\setcounter{secnumdepth}{3} 

\maketitle
\begin{abstract}
Two-dimensional cutting and packing problems aim to arrange pieces on a larger surface without overlap while minimizing material waste. Although extensively studied in industries such as textiles, traditional nesting research assumes piece geometries are fixed. In this work, we make the garment design itself a variable of the nesting optimization: we introduce a new formulation of the sewing pattern nesting problem in which panel geometry is modified under domain-aware constraints, preserving fit while allowing appearance to vary within designer-set bounds. We present an evolutionary nesting framework that embeds these modifications directly into a genetic algorithm (GA). Two novel mutation operators are proposed: one enables splitting panels into stitchable sub-panels, and the other adjusts garment design while preserving fit. Our method is evaluated on existing irregular nesting benchmarks and a new benchmark derived from a comprehensive dataset of sewing patterns. Our experiments demonstrate that evolutionary algorithms enhanced with the proposed garment-aware mutations achieve markedly better results than versions restricted to standard permutation-based operators, demonstrating the value of integrating design flexibility into nesting optimization. Our implementation is available at \url{https://github.com/fittedgenes/GarmentCode-Nesting}.

\ccsdesc[500]{Computing methodologies~Search methodologies}
\ccsdesc[300]{Computing methodologies~Genetic algorithms}
\ccsdesc[300]{Applied computing~Operations research}

\printccsdesc
\end{abstract}

\begin{keywords}
Irregular nesting, Computational garments, Evolutionary optimization
\end{keywords}

\section{Introduction}
\label{sec:intro}

Fabric waste is a significant and well-documented problem in garment production, with
cutting-stage losses consistently estimated at 10--30\% of total material
\cite{akter2022, cao2022, deSilva2023}. Minimizing this waste is the central concern
of two-dimensional irregular nesting: given the panels of a sewing pattern, the goal is to arrange them on a fixed-width fabric strip without overlap, minimizing the length of material consumed.

In this work, we use the following definition of a sewing pattern: it is a set of 2D panels --- flat fabric pieces --- that, when cut and assembled, form a three-dimensional garment. Each panel is a planar region bounded by straight edges and smooth curves. Despite decades of research \cite{bennell2008, guo2022}, all existing nesting formulations share a common assumption: panel geometry is fixed. The problem is treated as purely downstream --- given a set of shapes, find the best layout. In practice, however, garment panels are not immutable. A panel may be split into stitchable sub-panels, and design parameters such as length, flare, or neckline depth may be adjusted without compromising fit. These modifications are routinely made by sewists to reduce waste, yet they have never been incorporated into nesting optimization.

We introduce shape-modifiable nesting, a generalization in which panel geometry is a variable of the optimization rather than a fixed input. Building on the GarmentCode parametric pattern framework~\cite{korosteleva2023}, we propose two domain-aware modification operators: panel splits that introduce stitchable seams, and parametric adjustments that alter panel geometry while preserving fit.

Throughout, we hold to an explicit contract: \emph{fit is preserved, appearance is deliberately variable within designer-set bounds}. Body measurements never enter the search space, so every layout we return fits the same wearer, as a variant of the input design rather than a reproduction of it.

We embed these operators into a genetic algorithm alongside a novel variable-length crossover operator. This formulation also requires rethinking the objective; we propose a generalized policy-based objective that accommodates different notions of usable fabric. We evaluate our method on 98 patterns
from the GarmentCodeData dataset~\cite{korosteleva2024}, which we propose as a new benchmark for garment nesting research, as well as on the standard ESICUP~\cite{esicup2025} instances. We find that on this problem naive evolution without offspring outperforms full genetic algorithms, implying a rugged loss-landscape.

\section{Related Work}
\paragraph*{Irregular nesting.}
Two-dimensional irregular nesting has been studied extensively in operations
research~\cite{bennell2008, guo2022}. Constructive heuristics based on
bottom-left placement~\cite{chazelle1983} and no-fit polygons~\cite{art1966}
remain the dominant decoding layer, with TOPOS~\cite{oliveira2000topos}
establishing the placement scoring policies, one of which we adopt here.
Metaheuristic sequence optimizers — genetic algorithms~\cite{
jakobs1996}, simulated annealing~\cite{dowsland1995}, and tabu
search~\cite{blazewicz1993} — are typically layered on top, with GAs particularly suited to permutation-based sequencing.
At the local search level, Gomes and Oliveira~\shortcite{gomes2002} proposed a
2-exchange heuristic over panel permutations, iteratively swapping pairs to
improve layout quality; we use this as a direct baseline.
More recently, matheuristic hybrids~\cite{fang2024, liu2023} and exact
MIP formulations~\cite{lastra2024} have been proposed, though the latter
remain limited to small instances with translation-only placement.
A common limitation across all of these approaches is the assumption that
panel geometry is fixed: nesting is treated as a purely downstream task given
an immutable set of shapes. We challenge this assumption in our work.

\paragraph*{Sustainable garment design.}
Work on computational sustainability in fashion has largely remained separate
from nesting research. Zero-waste CAD systems~\cite{zhang2024}
enforce waste-free constraints through linked pattern edits, while
Bigger~\shortcite{bigger2021} uses a GA to optimise design parameters for circularity,
measuring fitness by total panel surface area rather than layout efficiency.
Qi~\shortcite{qi2025} formulates garment upcycling as a discrete-continuous
optimization problem, maximizing seam reuse when reassembling a source
garment. Niinim{\"a}ki~\shortcite{niinimaki2022} surveys the disconnect between zero-waste
fashion design and C\&P optimization, noting that the two traditions have yet
to be integrated. Our work directly addresses this gap by embedding
design modifications into the nesting loop itself.

\paragraph*{Benchmarks.}
Empirical evaluation of irregular nesting relies almost entirely on the
19 ESICUP instances~\cite{esicup2025}, of which only 7 correspond to garments
— too few to capture the geometric diversity of real sewing patterns.
We propose to use a curated subset of GarmentCodeData~\cite{korosteleva2024}, a large-scale dataset of parametrically generated sewing patterns, as a richer benchmark for
garment nesting research.

\section{Problem Formulation} \label{sec:problemdef}

\subsection{Notation and Setup}

Let $\mathcal{P} = \{p_1, \ldots, p_n\}$ be a set of $n$ garment panels — planar pieces with  potentially curved boundaries — to be translated and rotated by a set of allowed angles $\mathcal{O}_i$ without overlap on a rectangular fabric strip $R$ of fixed width $W$ and unbounded length. Prior to nesting, each panel is expanded uniformly to incorporate a global seam allowance and then linearized by replacement with an approximating polygon.

Let $T = \bigcup_{i=1}^n p_i$ denote the union of all placed panels $p_i \subseteq R$, which are required to be pairwise non-overlapping and fully contained within $R$. In standard 2D irregular strip packing, the objective is to minimize the occupied strip length $F = \max_x T - \min_x T$. This formulation treats panel geometry as immutable and assumes all fabric to the left of the rightmost placed panel constitutes the used region. This makes minimizing $F$ equivalent to maximizing the fill rate $\eta = \text{Area}(T) / (W \cdot F)$. We depart from both assumptions.

\subsection{Shape-Modifiable Nesting}

We introduce a generalization in which panels may be modified prior to placement, subject to domain-aware feasibility constraints. Formally, each panel $p_i$ allows a set of \emph{admissible modifications} $\mathcal{M}_i$. Applying $M_i \in \mathcal{M}_i$ to $p_i$ yields a modified panel set $P'_i = M_i(p_i)$, which may contain one or more sub-panels. Two modification types are considered:

\begin{itemize}
    \item \textbf{Splits.} A panel $p_i$ is divided along a cut line into two sub-panels $\{p'_{i,1}, p'_{i,2}\}$, subject to construction constraints --- cuts must not bisect darts, and skirt cuts run parallel to the hemline normal. A dart-aware split example (pant back panels) is provided in the supplementary material. The sub-panels are placed independently and later stitched to reconstruct $p_i$; the reconstructed panel has the same outline as $p_i$, so a split changes where seams fall in the finished garment but not its silhouette. A sub-panel produced by a split is not itself a candidate for further splitting (\cref{sec:mutation}).
    \item \textbf{Parametric adjustments.} GarmentCode decouples \emph{body parameters} — measurements such as height, waist, and hip circumference that encode the wearer's fit — from \emph{design parameters} — stylistic choices such as skirt flare, hem length, or neckline depth that govern panel geometry without affecting how the garment fits the body. Holding body parameters fixed while varying design parameters within user-specified bounds therefore changes panel shape in a fit-preserving way, without introducing additional seams. In this case, $P'_i = \{p'_i\}$ is a singleton set containing the reshaped panel.
\end{itemize}

The fit/appearance contract of \cref{sec:intro} is thus enforced structurally: splits leave fit and silhouette untouched and change only seam placement, while parametric adjustments leave fit untouched but do change appearance by construction --- a skirt with reduced flare is visibly a narrower skirt. The admissible set $\mathcal{M}_i$ is therefore not a fixed property of a panel but a designer-authored statement of how far the garment may drift from its input; the bounds used in our experiments are given in \cref{sec:mutation}.

\begin{figure}[H]
    \centering
    \includegraphics[width=0.9\linewidth]{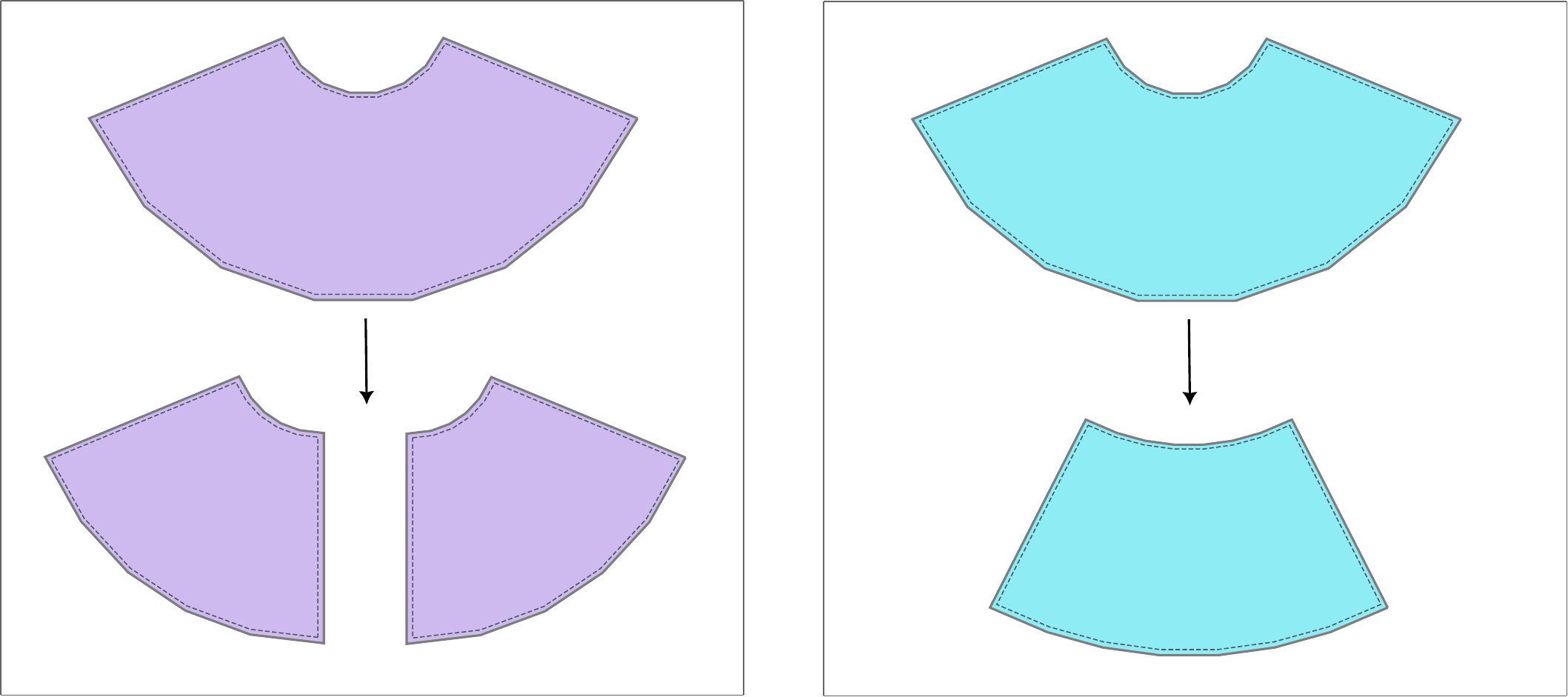}
    \caption{Examples of the two proposed panel modifications: \textbf{splits} (left) and \textbf{parametric adjustments} -- flare parameter adjusted from 0.75 (top-right) to 0.25 (bottom-right).}
    \label{fig:modifications}
\end{figure}

The placed region is then $T = \bigcup_i \bigcup_{p' \in P'_i} p'$, and the full constraint set is:
\begin{align}
    p'_{i,k} \cap p'_{j,\ell} &= \emptyset \quad \forall\, (i,k) \neq (j,\ell), \\
    p'_{i,k} &\subseteq R \quad \forall\, i, k, \\
    \theta_{i,k} &\in \mathcal{O}_i \quad \forall\, i, k, \\
    P'_i &= M_i(p_i),\quad M_i \in \mathcal{M}_i \quad \forall\, i,
\end{align}
where $p'_{i,k}$ denotes the $k$-th sub-panel of modified panel $i$ at orientation $\theta_{i,k}$.
Constraints~(1)--(2) enforce non-overlap and containment; (3) restricts rotations to grain-line-compatible orientations  $\mathcal{O}_i \subseteq \{0^\circ, 180^\circ\}$ permitted for panel $p_i$; (4) couples layout and shape decisions. Example modifications are shown in \cref{fig:modifications}.

\subsection{Policy-Based Objectives}\label{subsec:fitness}

The second generalization concerns what region of the strip counts as \emph{used}. In classical strip packing, the used region is implicitly the full-width rectangle $[\min_x T,\, \max_x T] \times [0, W]$, so minimizing $F$ and maximizing $\eta$ are interchangeable. This reflects industrial marker making, where a single guillotine cut terminates the layout. In other sewing contexts like home sewing, where fabric is purchased in finite quantities, the relevant waste is determined by the geometry of the placed pieces rather than a cut line. Irregularly shaped offcuts within the layout boundary may remain reusable, and the strip formulation systematically misclassifies them as waste.

We instead define the used region via a \emph{policy} $\pi$, a mapping $\pi(T, R) = U$ satisfying $T\subseteq~U\subseteq~R$. Three natural choices of policy $\pi$ arise, each inducing a
corresponding utilization $\eta_\pi = \frac{\mathrm{Area}(T)}{\mathrm{Area}(U_\pi)}$. The \emph{strip} policy sets $U = [\min_x T,\, \max_x T] \times [0, W]$, yielding $\eta_{\mathcal{S}} = \mathrm{Area}(T)/(F \cdot W)$. The \emph{bounding box} policy sets $U = \mathcal{B}(T)$, the axis-aligned enclosure of all placed pieces, yielding $\eta_{\mathcal{B}} = \mathrm{Area}(T)/\mathrm{Area}(\mathcal{B}(T))$. The \emph{concave hull} policy sets
$U = \mathcal{H}_\alpha(T)$, a tight non-convex enclosure that
excludes boundary voids from the used region, yielding $\eta_{\mathcal{H}} = \mathrm{Area}(T)/\mathrm{Area}(\mathcal{H}_\alpha(T))$.

The strip policy reduces $\eta_\pi$ to the classical fill rate. The bounding box and concave hull policies yield strictly tighter estimates of effective waste, excluding contiguous offcuts that remain reusable.

\subsection{Generalized Objective}
In standard strip packing, $\mathrm{Area}(T)$ is constant and $U$ is always the full-width strip up to $F$, so minimizing $F$ and maximizing
$\eta_{\mathcal{S}}$ are equivalent. Our formulation breaks this equivalence in two ways. First,
replacing the strip with a tighter policy $\pi$ decouples the two metrics even under fixed geometry,
since the used area now depends on the shape of the layout rather than solely
on $F$. Second, shape modifications cause $\mathrm{Area}(T)$ itself to vary,
so the two objectives can diverge. Therefore, we adopt a
combined objective that captures both dimensions:
\begin{equation}
   \Phi = \eta_{\pi} + \frac{L_{\max} - F}{L_{\max}}
\end{equation}
where $\eta_{\pi}$ is the fill rate under a chosen placement policy $\pi$ and
$F$ is the occupied strip length.
As a finite container length is required for computation, we bound the strip
by $L_{\max}$, chosen large enough that it does not act as a binding constraint
in practice. Since $L_{\max}$ is not an intrinsic problem quantity, the term $1/L_{\max}$ acts as an implicit weight on $F$ relative to $\eta_{\pi}$; in our experiments we set $L_{\max} = 800$~cm.
The generalized problem is then:
\begin{equation}
    \max_{M_1, \ldots, M_n,\; \theta,\; \sigma} \;\Phi \quad \text{subject to } (1)\text{--}(4),
\end{equation}
where $\sigma$ denotes the placement sequence and all feasibility constraints are enforced per sub-panel after modification.

\section{Method}
\label{sec:method}

We frame garment nesting as an evolutionary search over structured sewing pattern representations. Solutions are encoded as chromosomes in a genetic algorithm, where genes are garment panels carrying shape, rotation, and placement order. Each chromosome also carries GarmentCode design parameters, enabling the search to explore geometric variations alongside spatial rearrangements. The initial population is seeded with randomly generated chromosomes. In each generation, an elite pool of parents produces new individuals through two non-deterministic mechanisms: \textit{mutations}, where a single parent's chromosomes undergo a random change (\cref{sec:mutation}), and \textit{crossover}, where a pair of parents' genes are recombined (\cref{sec:crossover}). We additionally consider a simpler variant without crossovers: Naive Evolution (NE) that produces all non-elite offspring through mutation alone. To maintain genetic diversity, a fraction of randomly generated chromosomes is also introduced in each generation, in both evolution modes.

\begin{figure}[t]
    \centering
    \includegraphics[width=\linewidth]{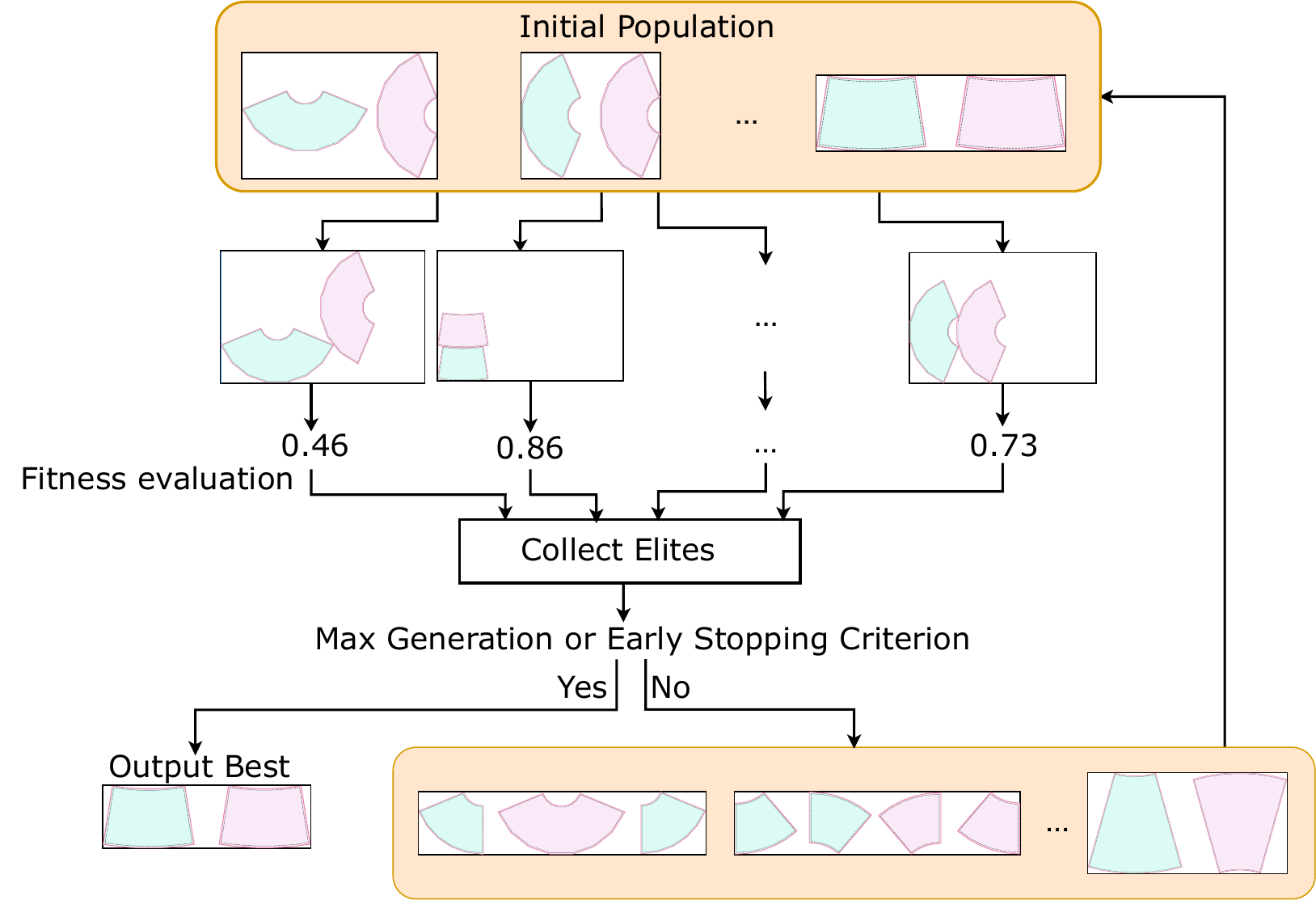}
    \caption{Overview of the evolutionary nesting pipeline.}
    \label{fig:gaflow}
\end{figure}

Fitness values of the child population are computed by decoding the chromosomes into 2D layouts using the constructive heuristics described in \cref{sec:decoders}. This involves extracting spatial descriptors (piece areas, bounding box, concave hull, and strip length).  The highest-fitness individuals from the new population and the elite parent pool are then chosen as the next generation's elite parents. After a specified number of generations, the highest-fitness individual is selected as the final solution. The full evolutionary loop is detailed in \cref{sec:evolutionary-loop}, and the pipeline is visualized in \cref{fig:gaflow}. We choose the bounding box policy for fill rate evaluation in the fitness function: it is inexpensive to compute and corresponds most closely to the classical notion of fill rate (a rectangular used region), while concave-hull evaluation is substantially more computationally expensive. Experiments additionally indicate high level of correlation between these two options (\cref{paragraph:correlation-analysis}). Accordingly, individuals are evaluated with $\Phi = \eta_{\mathcal{B}} + (L_{\max}-F)/L_{\max}$ as defined in \cref{sec:problemdef}; we still log $\eta_{\mathcal{H}}$ for analysis.

\subsection{Decoder} \label{sec:decoders}
The decoder transforms a chromosome into a concrete 2D layout within a rectangular
fabric container. We implement two constructive decoders: the Bottom-Left
heuristic~\cite{chazelle1983} and an NFP-based decoder following~\cite{oliveira2000topos},
which places each piece at the highest-scoring position along the evolving exterior
contour. Additional decoder details (including the NFP contour construction and scoring-policy comparison) are provided in Supplementary Appendix~A.

We use NFP with minimum $\mathrm{Area}(\mathcal{B})$ as the placement scoring policy, as it achieves the highest $\eta_{\mathcal{B}}$ across all configurations evaluated on 1{,}700 GarmentCodeData patterns; further policy comparison and definitions are provided in Supplementary Appendix~B.

\subsection{Crossover} \label{sec:crossover}

A chromosome is a permutation of the panel set, where each panel has a unique identifier. Standard crossovers (Partially Mapped Crossover (PMX)~\cite{goldberg1985}, Order Crossover (OX)~\cite{davis1985}) are insufficient because chromosomes carry two additional constraints:

\paragraph*{Split Integrity.} For every root panel $r$ with subpanel set $\mathcal{S}(r)$, the child must inherit either the unsplit root or the complete subpanel set from one parent:
\[
C \cap \mathcal{S}(r) \;\in\; \bigl\{ \{r\},\; \mathcal{S}(r) \cap P_1,\; \mathcal{S}(r) \cap P_2 \bigr\}.
\]

\paragraph*{Design Parameter Consistency.} A design parameter $d$ affects a
set of panels $\mathcal{P}_d$ that share seams whose geometry depends on $d$.
If the two parents assign different values to $d$, mixing panels from
$\mathcal{P}_d$ across parents would produce mismatched seam geometries.
Therefore, all panels in $\mathcal{P}_d$ must be inherited from the same parent. This constraint is transitive: if panels $p_i$ and $p_j$ are coupled by one
parameter, and $p_j$ and $p_k$ by another, then $p_i$, $p_j$, and $p_k$ must
all come from the same parent --- otherwise $p_j$ would have inconsistent
geometry with both neighbours. Formally, we build a conflict graph where
$(p_i, p_j)$ is an edge if the parents disagree on some parameter $d$ with
$p_i, p_j \in \mathcal{P}_d$, and require each connected component $K$ to be
inherited entirely from one parent:
\[
K \subseteq P_1 \quad \text{or} \quad K \subseteq P_2.
\]

We propose \emph{Cross-Stitch Crossover} (CSX), a novel OX variant satisfying
both constraints. $k$ contiguous segments are selected from Parent~1's root-order
sequence; every root therein (plus its consistency-closure) is assigned to
Parent~1 in an ownership map, the rest to Parent~2. The child is built by weaving
both parents' contributions in parallel under two clash-resolution modes:
\textbf{Sticky} (adopt the entire contiguous run from Parent~1 on a clash) and
\textbf{Lexicographic} (alternate one panel from each parent). A complementary
child is produced by flipping the ownership map. Full pseudocode and a worked
example are provided in the supplementary material.

\subsection{Mutation} \label{sec:mutation}

Mutations act on a single chromosome to introduce variation and escape local optima. Two classes are employed: standard permutation-based operators (swap, inversion, insertion, and scramble~\cite{davis1991}) and novel \emph{panel mutations}. Panel mutations modify panel geometry or composition, opening search space regions inaccessible to permutation operators. The simplest is \textbf{rotation}: $\textsc{Rotate}(p_i,\theta) = (p_i,\theta')$, $\theta' \in \mathcal{O}_i$. More advanced mutations exploit garment semantics.

Here inversion is the classical permutation operator~\cite{davis1991} --- it reverses a contiguous run of the placement sequence --- not a geometric reflection. We never mirror panels: a reflected piece has a reversed grain line and is not interchangeable with the original on directional or single-faced fabric. Rotations are limited to $\mathcal{O}_i \subseteq \{0^\circ, 180^\circ\}$ for the same reason.

\begin{figure}[h]
    \centering
    \setlength{\tabcolsep}{2pt}
    \renewcommand{\arraystretch}{1.1}
    \begin{tabular}{@{}
                    >{\centering\arraybackslash}m{1.2em}@{\hspace{2pt}}
                    >{\centering\arraybackslash}m{0.44\linewidth}
                    >{\centering\arraybackslash}m{0.44\linewidth}@{}}
        & \small Original & \small Split front \\[2pt]
        \rotatebox{90}{\small Flat pattern} &
        \includegraphics[width=\linewidth, trim=0cm 0cm 0cm 0cm, clip]{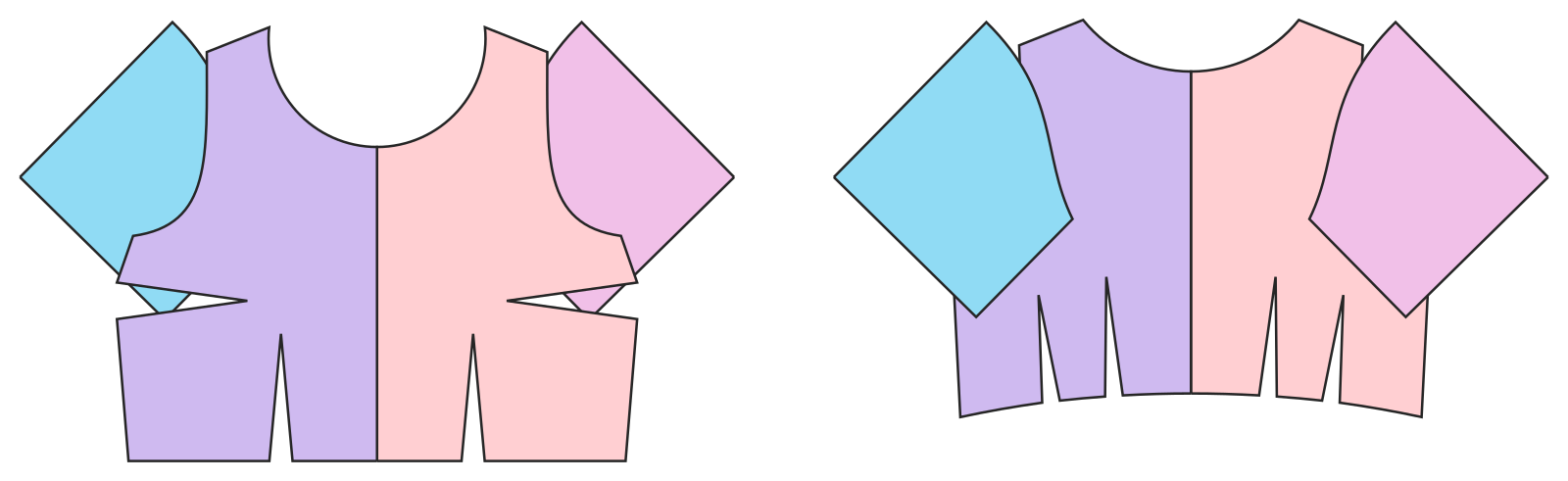} &
        \includegraphics[width=\linewidth, trim=0cm 0cm 0cm 0cm, clip]{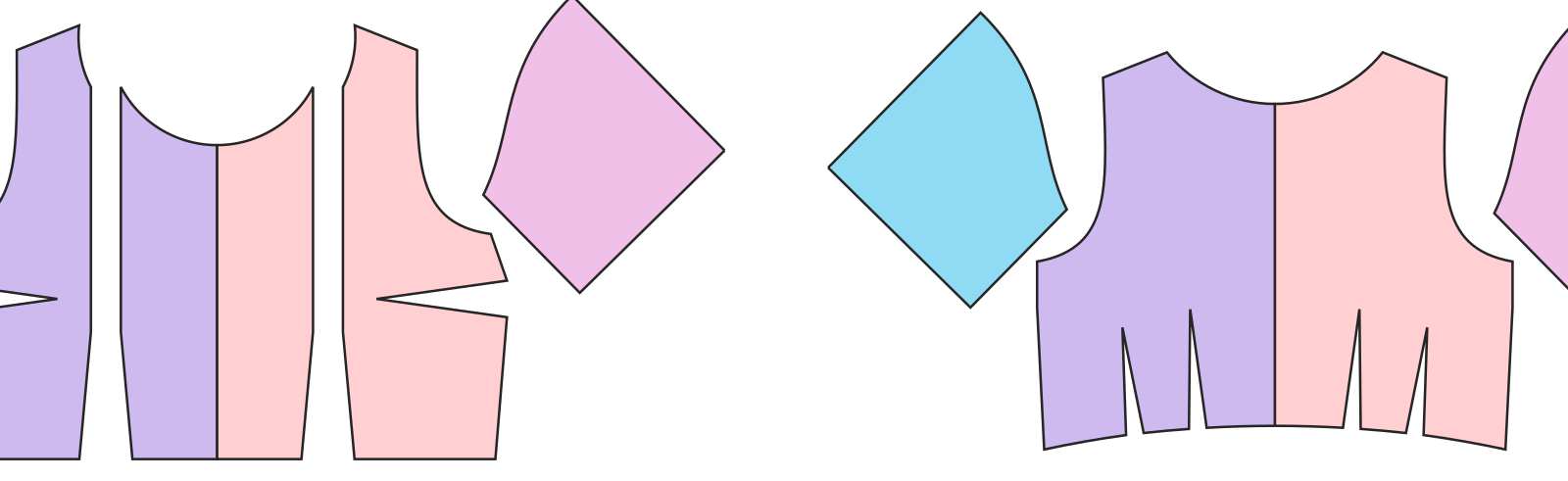} \\[4pt]
        \rotatebox{90}{\small 3D drape} &
        \includegraphics[width=\linewidth, trim=0cm 1.75cm 0cm 0cm, clip]{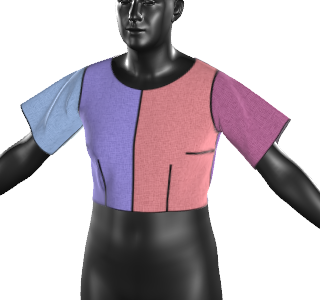} &
        \includegraphics[width=\linewidth, trim=0cm 1.5cm 0cm 0cm, clip]{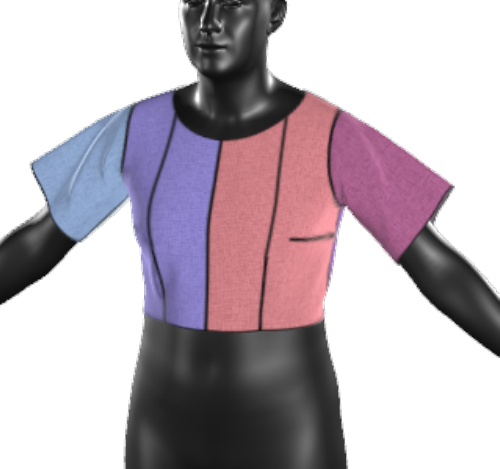}\\
    \end{tabular}
    \caption{Splitting the front bodice panels. A vertical style line through each
        waist-dart apex divides every front half into a centre-front and a side
        panel (right), turning the waist dart into a seam. Panels are
        colour-coded to their position on the body (cf.\ the 3D drape). The
        modification changes construction, not silhouette.}
    \label{fig:front-split}
\end{figure}

\paragraph*{Split Mutations.}
A panel is subdivided into two fragments stitched back in the finished garment,
allowing smaller pieces to fill gaps or reduce wasted space. An example split is shown in Fig. \ref{fig:front-split}. Cuts must respect
construction constraints, such as not truncating darts and keeping skirt cuts
parallel to the hemline normal; Supplementary Appendix~D gives the dart case in
detail. For polygon $P$, a vertical cut line at
$x_0 \in [x_{\min}(P),\, x_{\max}(P)]$ intersects $\partial P$ in candidate
interior segments $[q_{2k-1}, q_{2k}]$. The most area-balanced cut is selected:
\[
k^\star = \arg\min_{k} \left| \operatorname{area}(P_k^{(L)}) - \operatorname{area}(P_k^{(R)}) \right|,
\]
yielding $P^{(1)} = P_{k^\star}^{(L)}$ and $P^{(2)} = P_{k^\star}^{(R)}$.
In the parametric pipeline the cut point is placed at a proportion $p$ of the
arc length along the panel's waist edges (Supplementary Appendix~D), with $p$
sampled uniformly from $[0.3, 0.7]$ in all experiments (Supplementary Appendix~G). A sub-panel is never itself a candidate for
further splitting.
Splits are not free: measured over the 98 benchmark patterns, splitting a panel
adds a median of $48.5$\,cm$^2$ of fabric (mean $68.7$, IQR $21.1$--$85.9$,
$n = 1087$), the seam allowance along the new cut edge on each sub-panel. The
cost scales with the length of the cut, so the largest values come from
full-circle skirt panels. A qualitative illustration of the resulting
length--utilisation trade-off is provided in Supplementary Appendix~H, ``Splits and the Length--Utilisation Trade-off''.
Panel-specific construction constraints are handled by overriding the
cut-point selection for each garment class; we demonstrate this for pant
back panels as a proof of concept in Supplementary Appendix~D.

\begin{figure}[h]
    \centering
    \setlength{\tabcolsep}{2pt}
    \renewcommand{\arraystretch}{1.1}
    \begin{tabular}{@{}
                    >{\centering\arraybackslash}m{1.2em}@{\hspace{2pt}}
                    >{\centering\arraybackslash}m{0.30\linewidth}
                    >{\centering\arraybackslash}m{0.30\linewidth}
                    >{\centering\arraybackslash}m{0.30\linewidth}@{}}
        & \small $0.25$ (low) & \small $0.75$ (medium) & \small $1.52$ (high) \\[2pt]
        \rotatebox{90}{\small Flat pattern} &
        \includegraphics[width=\linewidth]{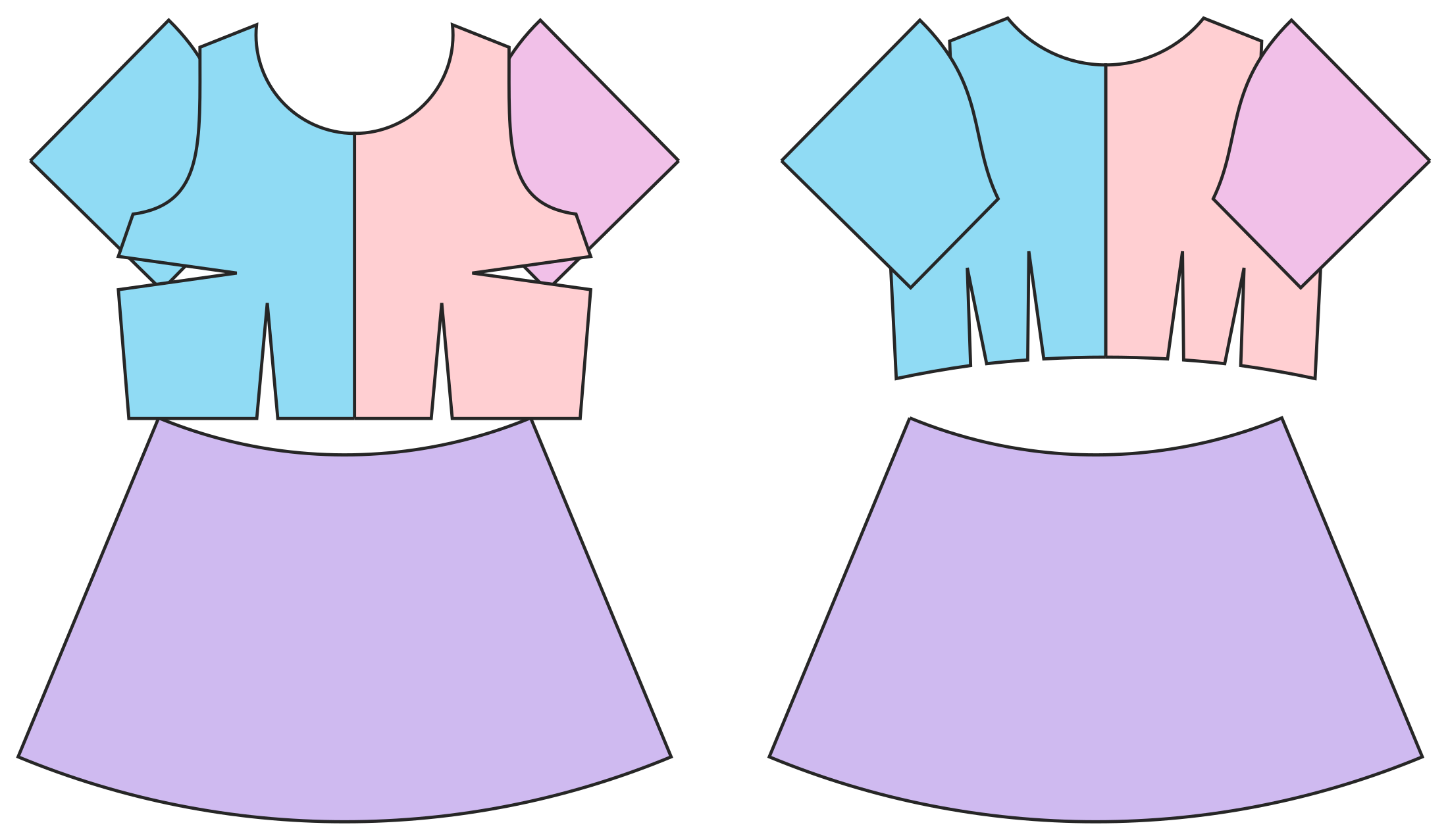} &
        \includegraphics[width=\linewidth]{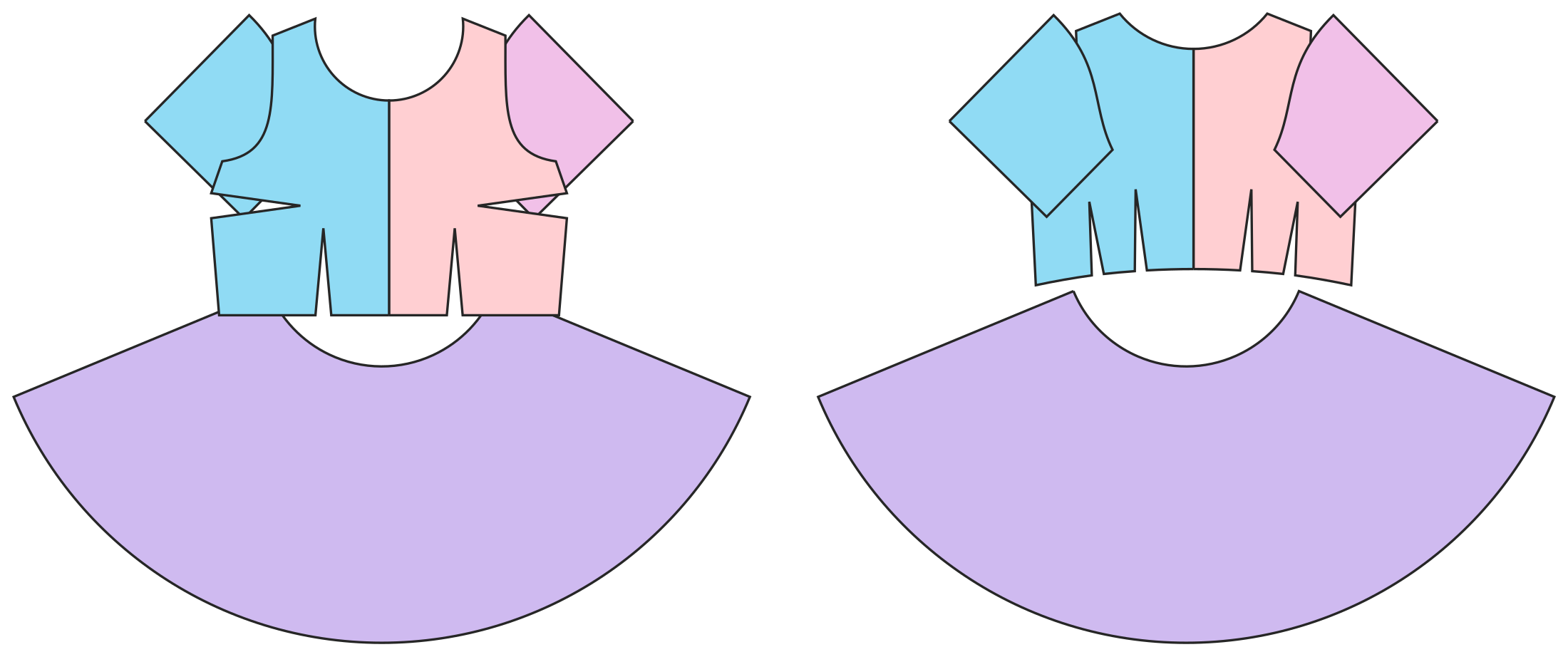} &
        \includegraphics[width=\linewidth]{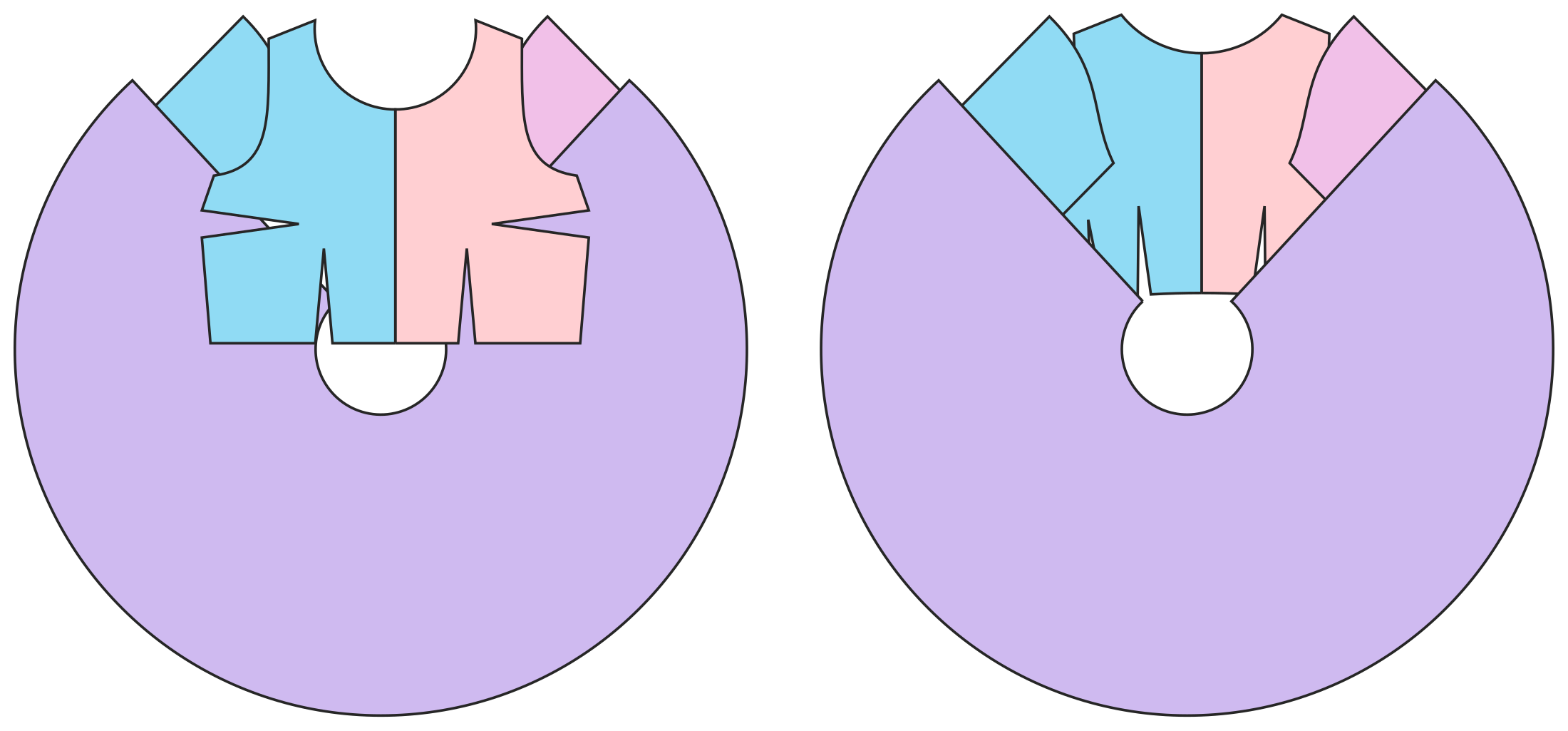} \\[4pt]
        \rotatebox{90}{\small 3D drape} &
        \includegraphics[width=\linewidth]{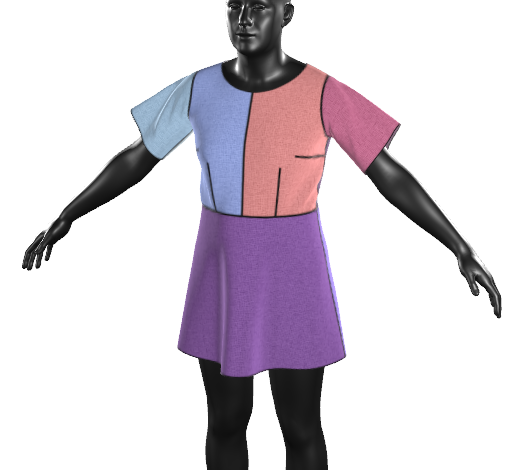} &
        \includegraphics[width=\linewidth]{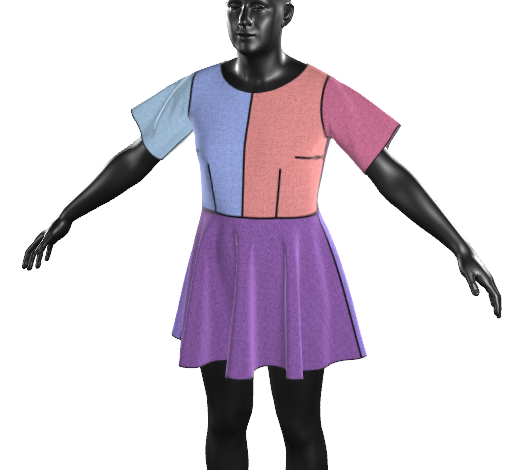} &
        \includegraphics[width=\linewidth]{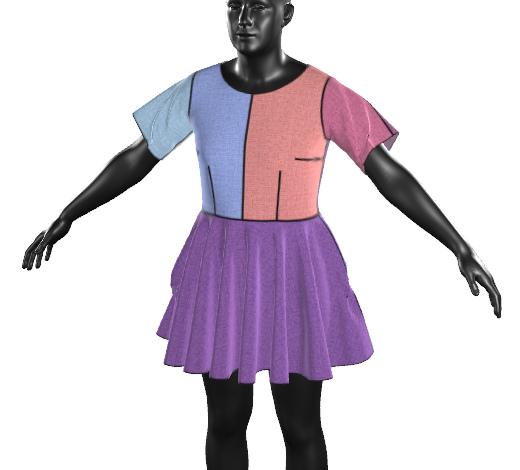} \\
    \end{tabular}
    \caption{%
        Effect of a single design-parameter mutation. Increasing the skirt
        flare parameter from $0.25$ to $0.75$ to
        $1.52$, with all other parameters held fixed, widens the flat skirt
        panel from a shallow arc toward a full circle (top) and produces a
        correspondingly fuller drape on the body (bottom).}
    \label{fig:suns-sweep}
\end{figure}

\paragraph*{Design Parameter Mutations.}
We exploit the body/design parameter decoupling of \textit{GarmentCode}
(\cref{sec:problemdef}) to mutate panel geometry without altering
garment fit: a design parameter is sampled within a user-defined margin of
its baseline value, and all panels whose geometry depends on it are
regenerated consistently, ensuring no seam mismatches are introduced. We set
this margin to $0.6$ in all experiments (Supplementary Appendix~G). Example design parameter mutations are shown in Fig. \ref{fig:suns-sweep}.

\subsection{Evolutionary Loop} \label{sec:evolutionary-loop}

\paragraph*{Initialization.} Panels are shuffled and assigned random rotations; each individual inherits the original design parameters; fitness is evaluated immediately.

\paragraph*{Evolution phases.} Each generation combines elites (carried over
unchanged), CSX offspring (produced from pairs of elites), mutants (elite
clones perturbed by the operators of \cref{sec:mutation}), and
randomly generated individuals for diversity. Population weights are given
in Supplementary Appendix~G.

\label{ch:results}

\setlength{\textfloatsep}{12pt plus 2pt minus 2pt}
\setlength{\floatsep}{10pt plus 2pt minus 2pt}
\setlength{\intextsep}{10pt plus 2pt minus 2pt}
\setlength{\dbltextfloatsep}{12pt plus 2pt minus 2pt}
\setlength{\dblfloatsep}{10pt plus 2pt minus 2pt}

\section{Evaluation}

We begin by examining the choice of objective, using a correlation analysis to motivate the combined objective $\Phi$ (\cref{paragraph:correlation-analysis}), and quantifying the contribution of each mutation operator (\cref{sec:effectiveness-mutation}). We then compare the full Genetic Algorithm against Naive Evolution and several baselines (\cref{tab:ga-vs-ne}), and ablate the two proposed operators to confirm both are necessary (\cref{tab:ablation}). Finally, we report results on the standard ESICUP garment instances (details in the supplementary material). All experiments are run on 98 patterns from GarmentCodeData and 7 ESICUP garment instances, repeated 5 times per pattern and averaged across runs. We report the metrics defined in \cref{subsec:fitness}.

\paragraph*{Runtime.}
A single run takes a median of 9 minutes (M3 MacBook Pro, 8 GB), with most runs between 3 and 45 minutes. Runtime is driven by panel count rather than generation count: small patterns (\textasciitilde 4 panels) finish in a couple of minutes, while 20-panel patterns take on the order of 45 minutes. This translates to roughly 30\,s per generation for a population of 100, i.e. \textasciitilde 0.3\,s per candidate layout, including NFP decode and fitness evaluation.

\subsection{Performance Analysis}

\subsubsection{Choice of Objective}
\label{paragraph:correlation-analysis}

\begin{figure}[h]
    \centering
    \includegraphics[width=0.9\linewidth]{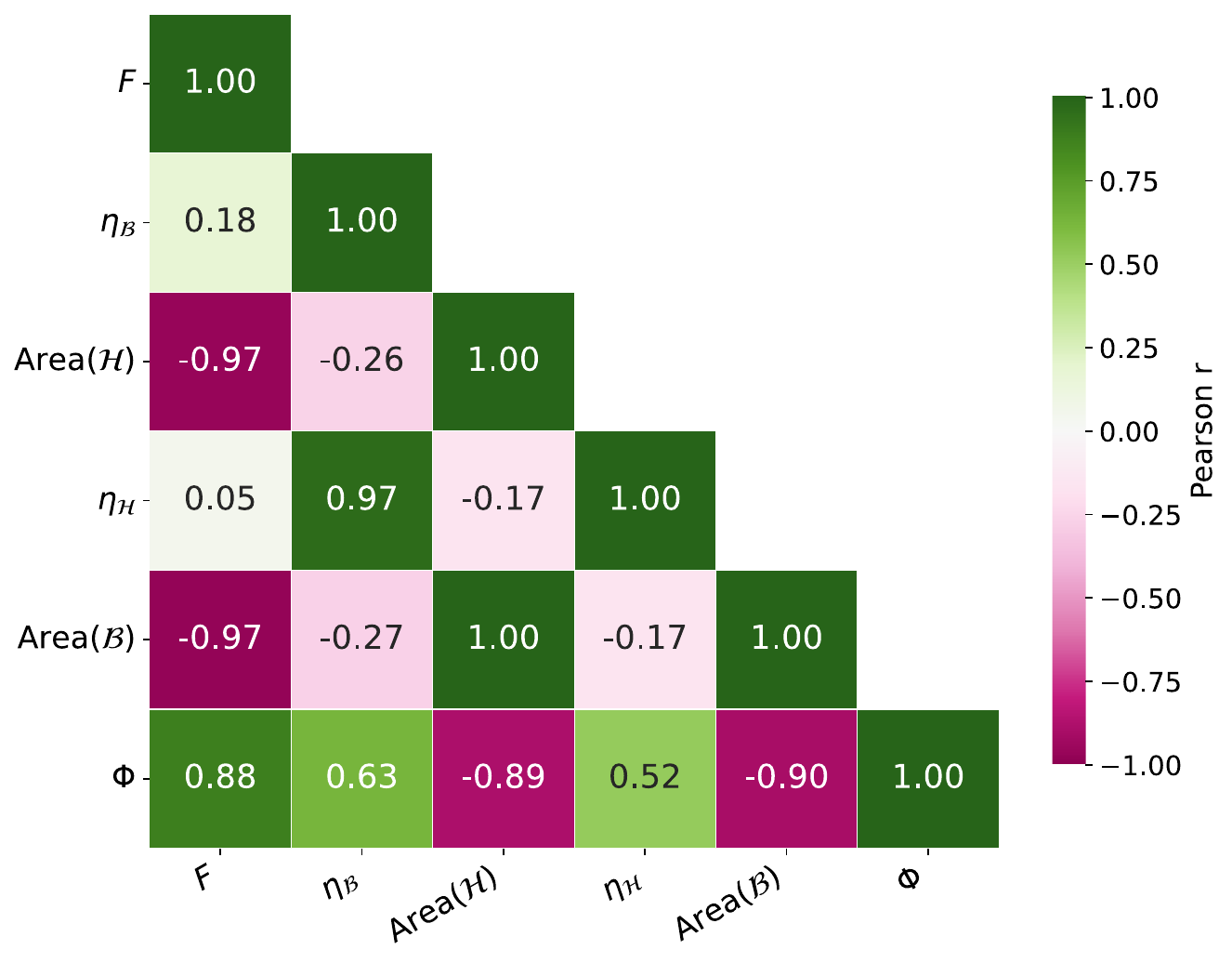}
    \vspace{-1.5em}
    \caption{Correlation matrix of objective functions.}
    \vspace{-1.5em}
    \label{fig:correlations}
\end{figure}

\cref{fig:correlations} shows the pairwise correlation matrix of all available metrics, evaluated on our chosen experimental setup, which optimizes the combined objective $\Phi$. 
Area-based objectives are strongly correlated with each other, as are utilization objectives.
Notably, $\eta_{\mathcal{B}}$ and $F$ are only weakly correlated in our generalized
setting, where piece geometries may change under constraints unlike in standard strip
packing. This confirms the need for a combined objective $\Phi = \eta_{\mathcal{B}} + (L_{\max} - F)/L_{\max}$, which shows moderate-to-strong correlation with all other objectives.
$\eta_{\mathcal{H}}$ and $\eta_{\mathcal{B}}$ are strongly correlated, suggesting utilization is a robust indicator regardless of hull definition. This confirms our choice of the bounding-box-based objective over the concave-hull one (\cref{sec:method}) and allows us to save on the computational cost of concave-hull evaluation.

\subsubsection{Effectiveness of Mutation Types}
\label{sec:effectiveness-mutation}

\begin{figure}[h]
    \centering
    \adjincludegraphics[trim={0 0 {.5\width} 0}, clip, width=\linewidth]{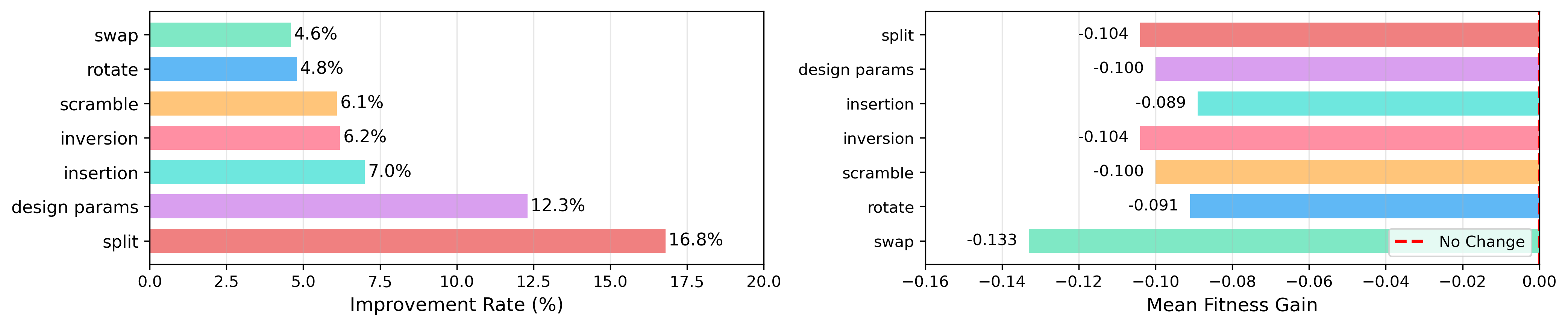}\\[6pt]
    \adjincludegraphics[trim={{.5\width} 0 0 0}, clip, width=\linewidth]{figures/Figure4_top_row.png}
    \caption{Mutation effectiveness (I): improvement rate per mutation type
    (top) and mean fitness impact across all mutants (bottom). Split and
    design parameter mutations succeed most often; swap has the most negative
    average impact.}
    \label{fig:mut_rates}
\end{figure}

\begin{figure}[h]
    \centering
    \adjincludegraphics[trim={0 0 {.5\width} 0}, clip, width=\linewidth]{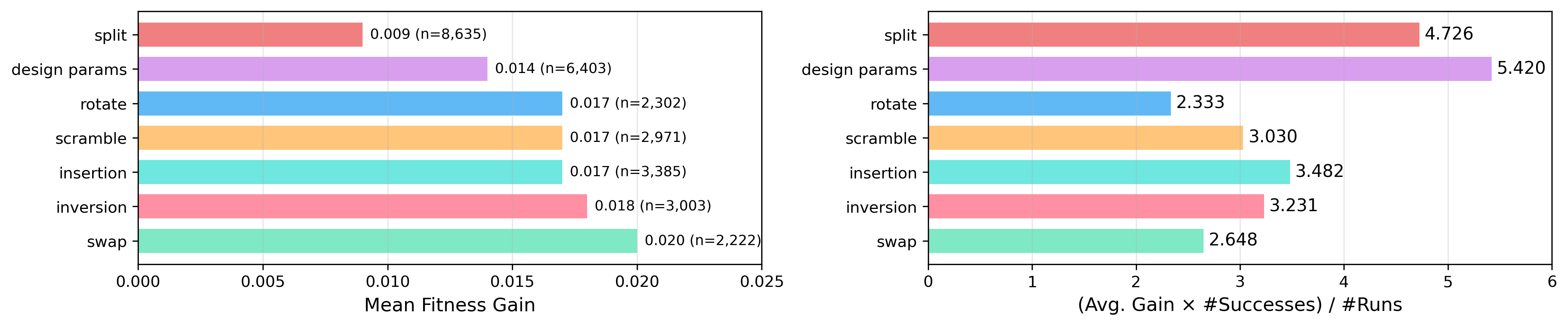}\\[6pt]
    \adjincludegraphics[trim={{.5\width} 0 0 0}, clip, width=\linewidth]{figures/Figure4_bottom_row.png}
    \caption{Mutation effectiveness (II): mean fitness gain conditional on
    success (top) and frequency-weighted gain per run (bottom). Despite
    smaller per-success gains, design parameter and split mutations dominate
    total progress due to their higher success rates.}
    \label{fig:mut_gain}
\end{figure}

We assess mutation effectiveness via the improvement rate (fraction of mutants
outperforming their elite parent) and the frequency-weighted gain per run
(\crefrange{fig:mut_rates}{fig:mut_gain}) across the NE runs with all mutation types enabled. Although swap and inversion yield the largest per-success gains, their low success rates (under 5\%) make their overall contribution small. By contrast, our proposed design parameter and split mutations achieve the highest success rates (12.3\% and 16.8\%) and contribute
most to total progress despite smaller individual gains. Per-type temporal mutation analysis is provided in Supplementary Appendix~E.


\subsection{Genetic Algorithm vs.\ Naive Evolution}

\cref{tab:ga-vs-ne} compares both crossover modes of the GA (sticky, lexicographic)
against NE with all mutations enabled, NE without rotations, 2-Exchange (2E), and Random
Search (RS). Qualitative case studies on rotation mutations, including when their removal becomes disqualifying, are provided in Supplementary Appendix~H, ``Behavior of Rotation Mutations''. Each generation in the GA consists of mutants produced by a single elite parent undergoing a mutation as well as offspring produced by two elite parents, whose genes are recombined through the crossover algorithm. NE, in contrast, does not produce offspring through crossover; all children of elite parents are mutants generated by a single parent.

The two GA crossover modes perform near-identically; sticky achieves slightly higher
$\eta_{\mathcal{H}}$ and lexicographic slightly shorter solutions. NE outperforms both GA variants on most metrics, particularly solution length (137.27\,cm vs.\ 140-142\,cm) and $\eta_{\mathcal{B}}$, suggesting that crossover does not
consistently yield stronger offspring than mutation alone, likely due to the geometric consistency constraints imposed during crossover. This suggests that crossovers are redundant for the optimization process.


Both evolutionary methods substantially outperform the baselines. The two baselines are not cleanly ordered: 2-Exchange achieves better efficiencies and smaller areas than Random Search, but Random Search attains a lower F (152.69 vs. 164.27) and marginally higher $\Phi$, suggesting that structured local search improves the secondary metrics while remaining prone to poor local optima on the primary objective. Both baselines are dominated by NE on all six metrics.

\newcolumntype{L}{>{\centering\arraybackslash\hsize=.85\hsize}X}
\newcolumntype{W}{>{\centering\arraybackslash\hsize=1.3\hsize}X}

\begin{table}[h]
    \centering
    \small
    \caption{Comparison of various optimization methods. Units for $F$ and Area are cm and cm$^2$ respectively. Best values per metric are highlighted in bold.}
    \label{tab:ga-vs-ne}
    \begin{tabularx}{\linewidth}{l L L L L W W}
    \toprule
    Method & $F \downarrow$ & $\eta_{\mathcal{B}} \uparrow$ & $\Phi \uparrow$ & $\eta_{\mathcal{H}} \uparrow$ & $\mathrm{Area}(\mathcal{B}) \downarrow$  & $\mathrm{Area}(\mathcal{H}) \downarrow$ \\
    \midrule
    GA (sticky)        & 141.43 & 80.34 & 1.6266 & 84.37 & 16439.4 & 15564.5 \\
    GA (lex.) & 140.33 & 80.24 & 1.6270 & 84.33          & 16439.5 & 15557.4 \\
    NE (full)          & \textbf{137.27} & \textbf{80.48} & \textbf{1.6332} & \textbf{84.39} & \textbf{16179.4} & \textbf{15336.9} \\
    2-Exchange         & 164.27 & 77.17 & 1.5664 & 82.59          & 17421.2 & 16191.3 \\
    Random Search      & 152.69 & 76.32 & 1.5723 & 82.05          & 17680.4 & 16307.4 \\
    \bottomrule
    \end{tabularx}
\end{table}

\subsection{Ablation Study}

\cref{tab:ablation} shows the effect of disabling the two proposed
operators — splits and design parameter mutations — individually and in
combination within NE.

\begin{table}[h]
    \centering
    \small
    \caption{Ablation study comparing NE configurations. Units for $F$ and Area are cm and cm$^2$ respectively. Best values per metric are highlighted in bold. Area$(\mathcal{B})$ and Area$(\mathcal{H})$ are not directly comparable across split configurations as we increase the total area by the seam allowance with each split.}
    \label{tab:ablation}
    \begin{tabularx}{\linewidth}{l L L L L W W}
    \toprule
    Method & $F \downarrow$ & $\eta_{\mathcal{B}} \uparrow$ & $\Phi \uparrow$ & $\eta_{\mathcal{H}} \uparrow$ & $\mathrm{Area}(\mathcal{B}) \downarrow $ & $\mathrm{Area}(\mathcal{H}) \downarrow $ \\
    \midrule
    NE (--s, --p) & 149.62 & 78.15 & 1.5945 & 83.22          & 17172.8 & 16036.0 \\
    NE (--splits)           & 138.33 & 80.09 & 1.6280 & 84.36 & \textbf{16081.6} & \textbf{15149.9} \\
    NE (--params)           & 146.96 & 79.53 & 1.6116 & 83.59          & 17025.0 & 16096.9 \\
    NE (--rotations)        & 138.99 & 79.95 & 1.6258 & 83.82          & 16298.2 & 15317.3 \\
    NE (full)               & \textbf{137.27} & \textbf{80.48} & \textbf{1.6332} & \textbf{84.39} & 16179.4 & 15336.9 \\
    \bottomrule
    \end{tabularx}
\end{table}

The full NE configuration achieves the best performance on $F$,
$\eta_\mathcal{B}$, $\Phi$, and $\eta_\mathcal{H}$, and disabling either
operator degrades all four, though their contributions are asymmetric:
design parameter mutations account for $9.69$\,cm of strip length against
$1.06$\,cm for splits. Splits nonetheless add a further $2.66$\,cm when
parameter mutations are unavailable ($149.62 \rightarrow 146.96$\,cm). At the
$140$\,cm width used here, splits save an average of $148.4$\,cm$^2$ of
fabric per pattern while adding an average of $106$\,cm$^2$ of seam
allowance across new cut edges.

\subsection{Evaluation on ESICUP}
For completeness, we evaluate on the 7 ESICUP garment instances (Albano,
Dagli, Mao, Marques, SHIRTS, SWIM, TROUSERS); results and discussion are
given in the supplementary material.

\subsection{Qualitative Results}
\begin{figure}[H]
    \centering
    \includegraphics[width=0.7\linewidth]{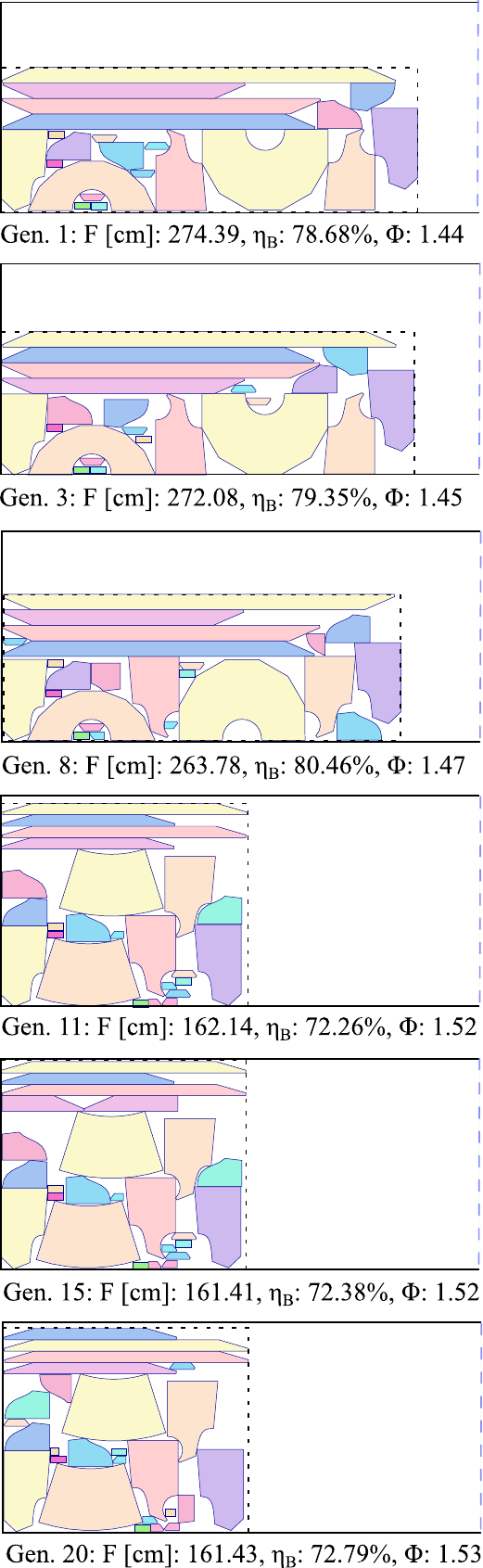}
    \caption{Design-parameter mutations in action. The flared, semicircular skirt panels of the first generations are reshaped into narrower panels that pack more tightly, while garment fit is preserved throughout. The black dashed line denotes the bounding box of the packed shapes; the blue dashed line marks where the $8$\,m fabric is truncated for display, as the region beyond the packed shapes is empty. Each panel reports the solution length $F$, the bounding-box fill rate $\eta_{\mathcal{B}}$, and the combined fitness $\Phi$ (fill rate plus normalized solution length).}
    \label{fig:example2}
\end{figure}

\Cref{fig:example2} shows how a representative layout evolves across
generations (stacked top to bottom, early to final); a second example appears as Fig.~H.8 in Supplementary Appendix~H. Several qualitative trends are visible. First, the used region (dashed) contracts steadily as the strip length $F$ shrinks: large panels settle into anchoring positions while smaller pieces and split fragments migrate into the concave voids between them, progressively eliminating the interior gaps that dominate the early layouts.
Second, the design-parameter mutation is plainly active --- in \cref{fig:example2}, the heavily flared, semicircular skirt panels of the initial generations are reshaped into narrower panels that pack far more tightly, all while preserving garment fit. Together these illustrate the mechanisms behind the quantitative gains reported above: the final layouts reach $\Phi = 1.53$ ($F = 161.43$\,cm, $\eta_{\mathcal{B}} = 72.79\%$) and
$\Phi = 1.56$ ($F = 134.78$\,cm, $\eta_{\mathcal{B}} = 73.22\%$) respectively.



\section{Conclusion}

We introduced \emph{domain-aware nesting}, a generalisation of 2D
irregular nesting in which panel geometry is an optimisation variable rather
than a fixed input. Concretely, we proposed two garment-aware mutation
operators — a \emph{split mutation} that introduces stitchable seams under
construction constraints and a \emph{design-parameter mutation} that
regenerates geometry within a fit-preserving margin via GarmentCode that extend our evolutionary optimization algorithm. We further introduced a policy-based objective generalising the classical strip-packing criterion, and proposed GarmentCodeData~\cite{korosteleva2024} as a new large-scale nesting benchmark.

Experiments confirm that both proposed operators are necessary and mutually complementary, together they improve fill rate by 2.33 percentage points and reduce solution length by 12.35 cm (8.25\%), and raise the combined objective $\Phi$ by 2.43\% over the permutation-only baseline. Metric correlation analysis validates the need for a combined objective in the shape-modifiable setting.
The algorithm exhibits graceful robustness: when rotations are withheld, design-parameter mutations compensate geometrically, recovering efficiency (see Supplementary Appendix~H, ``Behavior of Rotation Mutations'').

\paragraph*{Limitations and future work.}
Semantic mutations require GarmentCode-authored patterns; supporting arbitrary
inputs would require lifting them into a parametric representation. CSX does
not outperform mutation-only search, suggesting its consistency constraints
are overly conservative and that more adaptive recombination warrants
investigation. Further improvements — smarter initialization, compaction
post-processing, and adaptive mutation weighting — remain as future work.
The shape-modifiable paradigm also transfers naturally to additive
manufacturing and flat-pack design, where semantically guided splits and
parameter tuning could similarly expand the packing search space under
domain constraints.

\bibliographystyle{eg-alpha-doi}
\bibliography{bibliography}

\end{document}